\pdfoutput=1
\documentclass[11pt]{article}
\usepackage[dvipsnames]{xcolor}
\usepackage{graphicx}
\usepackage[]{acl}

\usepackage{times}
\usepackage{latexsym}

\usepackage[T1]{fontenc}
\usepackage[utf8]{inputenc}

\usepackage{microtype}

\usepackage{comment}
\usepackage{amsfonts}
\usepackage{amsmath}
\usepackage{siunitx}
\usepackage{booktabs}
\usepackage{tabularx}
\usepackage{multirow}
\usepackage{booktabs}
\usepackage[export, demo]{adjustbox}  
\usepackage{multicol, makecell}
\usepackage{bbm}
\usepackage{tablefootnote}
\usepackage{romannum}

\usepackage{tikz}
\newcommand*\circled[1]{\tikz[baseline=(char.base)]{
            \node[shape=circle,draw,inner sep=0.1pt] (char) {#1};}}

\newcolumntype{C}{>{\centering\arraybackslash}X}
\DeclareMathOperator*{\argmax}{argmax}
\definecolor{light-gray}{gray}{0.95}

\newcommand{\howard}[1]{\textcolor{red}{#1}}

\usepackage{pifont}% http://ctan.org/pkg/pifont
\newcommand{\cmark}{\ding{51}}%
\newcommand{\xmark}{\ding{55}}%

\definecolor{olive}{RGB}{0,153,51}

\title{Improving Hateful Meme Detection through Retrieval-Guided Contrastive Learning}

\author{Jingbiao Mei, Jinghong Chen, Weizhe Lin, Bill Byrne, Marcus Tomalin \\
Department of Engineering\\
University of Cambridge\\
Cambridge, United Kingdom, CB2 1PZ \\
  \texttt{\{jm2245, jc2124, wl356, wjb31, mt126\}@cam.ac.uk} \\}
\begin{document}
\pagenumbering{arabic}
\maketitle

\begin{abstract}
Hateful memes have emerged as a significant concern on the Internet.
Detecting hateful memes requires the system to jointly understand the visual and textual modalities. 
Our investigation reveals that the embedding space of existing CLIP-based systems lacks sensitivity to subtle differences in memes that are vital for correct hatefulness classification. We propose constructing a hatefulness-aware embedding space through retrieval-guided contrastive training. 
Our approach achieves state-of-the-art performance on the HatefulMemes dataset with an AUROC of 87.0, outperforming much larger fine-tuned large multimodal models.
We demonstrate a retrieval-based hateful memes detection system, which is capable of identifying hatefulness based on data unseen in training. This allows developers to update the hateful memes detection system by simply adding new examples without retraining — a desirable feature for real services in the constantly evolving landscape of hateful memes on the Internet.
\end{abstract}

 \textcolor{red}{This paper contains content for demonstration purposes that may be disturbing for some readers.}

\begin{comment}
    \begin{figure}[htbp!] 
\centering    
\includegraphics[width=0.49\textwidth]{Figs/Multimodal_Confounders_6memes.png}
\caption{Illustrative (not in the dataset) examples from \citealt{KielaFBHMC2020}. Memes on the left are mean, the ones in the middle are benign image confounders, and those on the right are benign text confounders.}
\label{fig:conf_HMC}
\end{figure}
\end{comment}

\section{Introduction}

\begin{figure}[htbp!] 
\small

\includegraphics[width=0.49\textwidth]{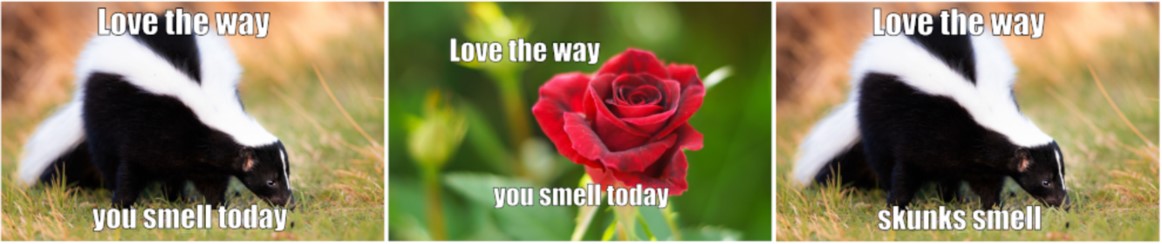}

\textit{Prediction}: Benign \textcolor{red}{\xmark} \hspace{2em} Benign \textcolor{green}{\cmark} \hspace{3.5em} Benign \textcolor{green}{\cmark}
\caption{Illustrative examples from \citealt{KielaFBHMC2020}. The meme on the left is hateful, the middle one is a benign image confounder, and the right one is a benign text confounder. We show HateCLIPper's \cite{KumarHateClip2022} \textit{prediction} below each meme. HateCLIPper misclassifies the hateful meme on the left as benign.}
\label{fig:conf_HMC}
\end{figure}

The growth of social media has been accompanied by a surge in hateful content. %A recent study shows that between 2019 and 2021, posts containing hate speech related to race or ethnicity were published on the Internet at an average rate of once every 1.7 seconds~\cite{ditch_the_label_2021}. 
Hateful memes, which consist of images accompanied by texts, are becoming a prominent form of online hate speech. This material can perpetuate stereotypes, incite discrimination, and even catalyse real-world violence. %To handle the large volume of potential hateful content and prevent viral circulation,
To provide users the option of not seeing it, hateful memes detection systems have garnered significant interest in the research community~\cite{KielaFBHMC2020, TamilTroll2020, suryawanshi-etal-2020-MultiOFF, pramanickCovidMeme2021, LiuFigMemes2022, HossainMUTEMeme2022, PrakashTotalDefMeme2023, SahinARCHateSpeechEvent2023}.
% However, automatically detecting hateful memes remains a formidable challenge due to their unique nature. Memes typically consist of both images and text, which requires models to comprehend the combined message of both two modalities.
% The challenge lies in that subtle differences in either image or text may lead to completely different meanings.
% For example, \citet{KielaFBHMC2020} introduced the concept of "confounder" memes. These confounder memes were created by altering either the image or the text from the original hateful memes, resulting in benign memes, as shown in Figure~\ref{fig:conf_HMC}.
% These examples demonstrate that an image or text can appear benign or hateful depending on subtle contextual cues found in either modality. This mirrors the behavior of real memes on the Internet, where the nuanced elements often contribute to their hateful nature.

Correctly detecting hateful memes remains difficult. Previous literature has identified a prominent challenge in classifying "confounder memes", in which subtle differences in either image or text may lead to a completely different meaning~\citep{KielaFBHMC2020}.  As shown in Figure \ref{fig:conf_HMC}, the top left and top middle memes share the same caption. However, one of them is hateful and the other benign depending on the accompanying images. Confounder memes resemble real memes on the Internet, where the combined message of images and texts contribute to their hateful nature. %Previous works attempted to tackle the challenge by leveraging outside knowledge to ground the reasoning \cite{RonHMC1st2020}, or building stronger multimodal fusions in the early stage ~\cite{PramanickMomenta2021} and intermediate stage ~\cite{KumarHateClip2022}.
Even state-of-the-art models, such as HateCLIPper~\cite{KumarHateClip2022}, exhibit limited sensitivity to nuanced hateful memes. 

We find that a key factor contributing to misclassification is that confounder memes are located in close proximity in the embedding space due to the similarity of text or image content.  For instance,  HateCLIPper's embedding of the confounder meme in Figure \ref{fig:conf_HMC} has a high cosine similarity score with the left anchor meme even though they have opposite meanings. This poses challenges for the classifier to distinguish harmful and benign memes. %, leading to suboptimal performance.

%To address this issue, 
We propose ``\textbf{Retrieval-Guided Contrastive Learning}'' (\textbf{RGCL}) to learn hatefulness-aware vision and language joint representations. We align the embeddings of same-class examples that are semantically similar with pseudo-gold positive examples and separate the embeddings of opposite-class examples with hard negative examples. We dynamically retrieve these examples during training and train with a contrastive objective in addition to cross-entropy loss. RGCL achieves higher performance than state-of-the-art large multimodal systems on the HatefulMemes dataset with far fewer model parameters. We demonstrate that the RGCL embedding space enables the use of K-nearest-neighbor majority voting classifier. The encoder trained on HarMeme \cite{pramanickCovidMeme2021} can be applied to HatefulMemes \cite{KielaFBHMC2020} without additional training while maintaining high AUC and accuracy using the KNN majority voting classifier, even outperforming large multi-modal models under similar settings. This allows efficient transfer and update of hateful memes detection systems to handle the fast-evolving landscape of hateful memes in real-life applications.
Our contributions are:
\begin{enumerate}
    \item We propose RGCL for hateful memes detection which learns a hatefulness-aware embedding space via an auxiliary contrastive objective with dynamically retrieved examples. We propose to leverage novel pseudo-gold positive examples to improve the quality of positive examples.
    
    \item Our proposed approach achieves state-of-the-art performance on HatefulMemes and the HarMeme. We show RGCL's capability across various domains of meme classification tasks on MultiOFF, Harm-P and Memotion7K. 
    % % \item a novel auxiliary task-specific loss based on similarity scores with pairs of retrieved hard negative, retrieved pseudo-gold positive, and in-batch negative examples to augment the cross-entropy loss. As a result, our system improves the model's ability to leverage subtle contextual cues presented in the meme and thus performs better on confounders compared to the previous state-of-the-art HateCLIPper. 
    % \item We achieved state-of-the-art performance in both the HatefulMemes dataset and Harmful Memes dataset, suggesting the system generalizes well to different domains of memes.  
    \item 
    %We demonstrate that the retrieval-based KNN majority voting classifier outperforms the zero-shot performance of large multimodal models of a much larger scale. This allows easy updating and extension of hateful meme detection systems without retraining. 
    Our retrieval-based KNN majority voting classifier facilitates straightforward updates and extensions of hateful meme detection systems across various domains without retraining. With RGCL training, the retrieval-based classifier demonstrates strong cross-dataset generalizability, making it suitable for real services in the dynamic environment of online hateful memes.
    % \item We demonstrate  a novel retrieval-based hateful meme prediction method that can effectively and efficiently transfer to unseen data that outperforms the zero-shot performance of large multimodal models of much larger scales. 
\end{enumerate}

\section{Related Work}
%\subsection{Hateful Memes Detection}
%The Hateful Memes Challenge competition \cite{KielaFBHMC2020} released a benchmark dataset for hateful meme detection. 
%The best baseline model from the challenge is the Visual BERT, achieving an AUROC of 75.4. The prize-winning solution \cite{RonHMC1st2020, VilioHMC2nd2020, RizaHMC3rd2020} of the challenge managed to push the AUROC to 84.5 with additional extracted features and ensembling of models. 
\textbf{Hateful Meme Detection Systems} in previous work can be categorized into three types:
% We categorise previous hateful meme detection systems into three types: 
Object Detector \textit{(OD)-based} vision and language models, \textit{CLIP}~\cite{clip2021} encoder-based systems, and Large Multimodal Models \textit{(LMM)}.

\textit{OD-based} models such as VisualBERT \cite{VisualBert2019}, OSCAR \cite{li2020oscar}, and UNITER \cite{Uniter2019} use Faster R-CNN \cite{fater_RCNN_2015} based object detectors~\cite{Anderson2017up-down, vinVL2021} as the vision model.
The use of such object detectors results in high inference latency \cite{Kim_ViLT2021}.%, making them less suitable for real-world hateful meme classification.

\textit{CLIP-based} systems have gained popularity for detecting hateful memes due to their simpler end-to-end architecture. %MOMENTA \cite{PramanickMomenta2021} and PromptHate \cite{caoPromptHate2022} augment CLIP representations with additional features such as text attributes and image captions.
% While these augmentations improve performance, they also introduce additional latency to the inference process.
HateCLIPper \cite{KumarHateClip2022} explored different types of modality interaction for CLIP vision and language representations to address challenging hateful memes. In this paper, we show that such CLIP-based models can achieve better performance with our proposed retrieval-guided contrastive learning. 

\textit{LMMs} like Flamingo~\cite{Flamingo22} and LENS~\cite{BerriosLens2023} have demonstrated their effectiveness in detecting hateful memes. Flamingo 80B achieves a state-of-the-art AUROC of 86.6, outperforming previous CLIP-based systems although requiring an expensive fine-tuning process. %However, in this paper, we demonstrate that CLIP-based model can achieve better performance than such LMMs with our proposed retrieval-guided contrastive learning.

\noindent\textbf{Contrastive Learning} is widely used in vision tasks \cite{Schroff_FaceNet_2015, Song_metriclearningLifetedFeatureEmbedding_2016, Harwood_SmartMiningDeepMetricLearning_2017, Suh_StochasticHardExampleMiningForDeepMetricLearning2019} and retrieval tasks , however, its application to multimodally pre-trained encoders for hateful memes has not been well-explored. 
\citet{LippeHMFramework2020} incorporated negative examples in contrastive learning for detecting hateful memes. However, due to the low quality of randomly sampled negative examples, they observed a degradation in performance. In contrast, our paper shows that by incorporating dynamically sampled positive and negative examples, the system is capable of learning a hatefulness-aware vision and language joint representation. 

\noindent\textbf{Sparse retrieval} methods, such as BM-25 \cite{Robertson_BM25_2009} have been used in contrastive learning to obtain collections of hard triplets \cite{dpr2020, Schroff_FaceNet_2015, Khattab_Zaharia_COLBERT_2020,Nguyen2023PassageBM25}. In contrast, \textbf{dense retrieval}, which is based on vector similarity scores, has been widely adopted for various passage retrieval tasks \cite{dpr2020, Santhanam_ColBERTV2_2021,Diaz2021PR, Herzig2021QATable, Lin2023FLMR, Lin2024Preflmr}. Our method leverages dense retrieval to dynamically select both hard negative and pseudo-gold positive examples.

\section{RGCL Methodology}
\begin{figure*}[hbpt]
    \centering
    \includegraphics[width=\textwidth]{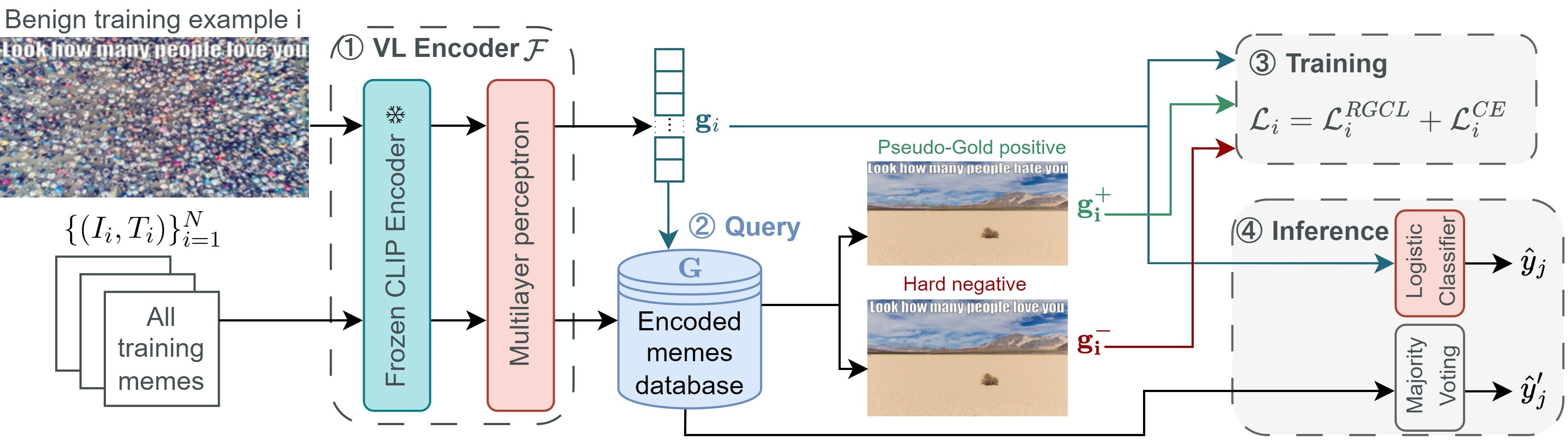}
    \caption{Model overview. \circled{1} Using VL Encoder $\mathcal{F}$ to extract the joint vision-language representation for a training example $i$. Additionally, the VL Encoder encodes the training memes into a retrieval database $\mathbf{G}$. \circled{2} During training, pseudo-gold and hard negative examples are obtained using the Faiss nearest neighbour search. During inference, $K$ nearest neighbours are obtained using the same querying process to perform the KNN-based inference. \circled{3} During training, we optimise the joint loss function $\mathcal{L}$. \circled{4} For inference, we use conventional logistic classifier and our proposed retrieval-based KNN majority voting. For a test meme $j$, we denote the prediction from logistic regression and KNN classifier as $\hat{y}_j$ and $\hat{y}'_j$, respectively. }
    \label{fig:Model_Architecture}
\end{figure*}

%\subsection{Feature Extraction}
In each training example
$\{ (I_i,T_i,y_i)\}_{i=1}^N$, %$\{ (I_i,T_i)\}_{i=1}^N$
$I_i \in \mathbb{R}^{C\times H \times W }$ is the image portion of the meme in pixels; $T_i$ is the caption overlaid on the meme; $y_i\in\{0,1\}$ is the meme label, where 0 stands for benign, 1 for hateful.

We leverage a Vision-Language (VL) encoder to extract image-text joint representations from the image and the overlaid caption: 
\begin{equation}
   \mathbf{g}_i = \mathcal{F}(I_i, T_i)
    \label{eq:Encoder}
\end{equation}
We encode the training set with our VL encoder to obtain the encoded retrieval vector database $\mathbf{G}$:
\begin{equation}
    \mathbf{G}=\{(\mathbf{g}_i, y_i)\}_{i=1}^N
    \label{eq:database}
\end{equation}
We index this retrieval database with Faiss \cite{johnson_Faiss2019billion} to perform training and retrieval-based KNN classification.

As shown in Figure~\ref{fig:Model_Architecture}, the VL encoder comprises a frozen CLIP encoder followed by a trainable multilayer perceptron (MLP). The frozen CLIP encoder encodes the text and image into embeddings that are then fused into a joint VL embedding before feeding into the MLP.

We use HateCLIPper \cite{KumarHateClip2022} as our frozen CLIP encoder. The model architecture is detailed in Appendix~\ref{appendix:hateclipper}. In Sec.\ref{sec:abl_encoder}, we compare different choices of the frozen CLIP encoder to demonstrate that our approach does not depend on any particular base model.

\subsection{Retrieval Guided Contrastive Learning}
%Retrieval-Guided Contrastive Learning aims to learn hatefulness-aware vision and language joint representations.
For each meme in the training set (the ``anchor meme''), we dynamically obtain three types of contrastive learning examples:  (1) pseudo-gold positive; (2) hard negative; (3) in-batch negative to train our proposed retrieval-guided contrastive loss.

(1) Pseudo-gold positive examples are same-label samples in the training set that have high similarity scores under the embedding space. Incorporating these examples pulls same-label memes with similar semantic meanings closer in the embedding space. 

% (1) Pseudo-gold positive examples are samples in the training set that share similarities with the anchor meme in the embedding space with the same labels.  
% By clustering memes with similar semantic meanings within the embedding space, we can strengthen the model's ability to capture a wide range of semantic relationships. 

(2) %Hard negative examples \cite{Schroff_FaceNet_2015} are samples in the training set that share similarities with the anchor meme in the embedding space but carry different labels.
Hard negative examples \cite{Schroff_FaceNet_2015} are opposite-label samples in the training set that have high similarity scores under the embedding space. These examples are often confounders of the anchor memes. By incorporating hard negative examples, we enhance the embedding space's ability to distinguish between confounder memes.
%We separate the embeddings of confounding examples by incorporating retrieved hard negative examples.
%Hard negative examples are memes that the current embedding space has failed to distinguish correctly, and are instances of misclassification or confounder cases in the dataset. Introducing training signals with hard negative examples can effectively enhance the embedding space's ability to distinguish between confounder memes.

(3) For a training sample $i$, the set of in-batch negative examples \cite{YihToutanovaPlattMeek2011inbatchnegative, Henderson2017effiNLPinbatchnegative} are the examples in the same batch that have a different label as the sample $i$.
In-batch negative examples introduce diverse gradient signals in the training and this causes the randomly selected in-batch negative memes to be pushed apart in the embedding space.

Next, we describe how we obtain these examples to train the system with Retrieval-Guided Contrastive Loss. %Finally we introduce the retrieval-based KNN majority voting.
%We dynamically retrieve these examples during training and train with the Retrieval-Guided Contrastive Loss in addition to cross-entropy loss.
\subsubsection{Finding pseudo-gold positive examples and hard negative examples}
%Pseudo-gold positive examples facilitate the clustering of memes within the same class that exhibit similar semantic characteristics, thereby strengthening the model's ability to capture a wide range of semantic relationships. In contrast, hard negative examples are samples in the training set that share similarities with the anchor meme in the embedding space but carry different labels. In essence, 
%We first encode the training set with our VL encoder. % %In other words, we retrieve examples based on the embedding space from the last training iteration. 
%The encoded retrieval vector database is denoted as $\mathbf{G}$:
%\begin{equation}
%    \mathbf{G}=\{(\mathbf{g}_i, y_i)\}_{i=1}^N
%    \label{eq:database}
%\end{equation}
%Due to the computational expense, 
%\howard{We update the database after every epoch.}
For a training sample $i$, we obtain the pseudo-gold positive example and hard negative example from the training set with Faiss nearest neighbour search \cite{johnson_Faiss2019billion} which computes the similarity scores between sample $i$'th embedding vector $\mathbf{g}_i$ and any target embedding vector $\mathbf{g}_j \in \mathbf{G}$. The encoded retrieval vector database $\mathbf{G}$ is updated after each epoch.

We denote the pseudo-gold positive example's embedding vector:
\begin{equation}
    \mathbf{g}_i^{+} = \argmax_{\mathbf{g}_j \in \mathbf{G} / \mathbf{g}_i,y_i=y_j}   \, \textrm{sim}(\mathbf{g}_i, \mathbf{g}_j),
\end{equation}
similarly for the hard negative example's embedding vector:
\begin{equation}
    \mathbf{g}_i^{-} = \argmax_{\mathbf{g}_j \in \mathbf{G},y_i \not=y_j} \, \textrm{sim}(\mathbf{g}_i, \mathbf{g}_j).
\end{equation}
We use cosine similarity for similarity measures.

%We found the best performance is achieved with $K^{hard}=1$, which is consistent with Dense passage Retrieval \cite{dpr2020}.
%Finally, we obtain the corresponding embedding vectors $\mathbf{G}_{i}^{--}$ with the current encoder state for the set of hard negative examples.
%\begin{equation}
%    \mathbf{G}_{i}^{--} = \{\textrm{Enc}(I,T)) \textrm{ \textbf{for} } (I,T,y) \in \mathcal{H}_i\}
%\end{equation}
%\subsubsection{In-batch negative examples}
%To enhance training stability and encourage robust learning,
 We denote the embedding vectors for the in-batch negative examples as $\{ \mathbf{g}_{i,1}^-, \mathbf{g}_{i,2}^-,...,\mathbf{g}_{i,n^{-}}^- \}$. 
We concatenate the hard negative example with the in-batch negative examples to form the set of negative examples $\mathbf{G}_{i}^{-}= \{ \mathbf{g}_{i}^-,\mathbf{g}_{i,1}^-, \mathbf{g}_{i,2}^-,...,\mathbf{g}_{i,n^{-}}^- \}$
%There are a total of $n^{-}$ in-batch negative examples correspond to the training sample $i$.
%A set of in-batch negative examples $\mathcal{N}_i$ for the $i$th sample triplet $(I_i,T_i, y_i)$ is defined as the set of examples in the same batch that have the opposite label $y_k$ as label $y_i$. 
%\begin{equation}
%\small
%\{(I_k,T_k,y_k): \,\,k=1,...,B \textrm{ \textbf{and} }  y_k \not= y_i  \} 
%\end{equation}
%We denote the number of in-batch negative examples for the triplet $(I_i,T_i, y_i)$ as $K_i^- = |\mathcal{N}_i|$.
%The embedding vectors of the in-batch negative examples are denoted as $\mathbf{G}_{i}^{-}$ which is obtained directly from the training step.   
\subsubsection{RGCL training and inference}
%$(\mathbf{g}_i, \mathbf{g}_i^{+}, \mathbf{g}_{i}^{--}, \mathbf{g}_{i,1}^-,...,\mathbf{g}_{i,n^{-}}^-)$is the vector representation of the original, pseudo-gold positive, hard negative, and in-batch negative examples corresponding to a training example $i$.
Following previous work \cite{KumarHateClip2022, KielaFBHMC2020, PramanickMomenta2021}, we use logistic regression to perform memes classification as shown in Figure~\ref{fig:Model_Architecture}. We denote the output from the logistic regression as $\hat{y}_j$ for sample $j$.

To train the logistic classifier and the MLP within the VL Encoder, we optimize a joint loss function. The loss function consists of our proposed Retrieval-Guided Contrastive Learning Loss (\textit{RGCLL}) and the conventional cross-entropy (\textit{CE}) loss for logistic regression:
\begin{align}
    \mathcal{L}_i &= \mathcal{L}_i^{RGCLL} + \mathcal{L}_i^{CE}\nonumber \\
            &= \mathcal{L}_i^{RGCLL} + (y_i\log \hat{y}_i + (1-y_i)\log(1-\hat{y}_i)),
\label{eq:RGCLL}
\end{align}
where the \textit{RGCLL} is computed as:
\begin{align}
\nonumber \mathcal{L}_i^{RGCLL}&= L(\mathbf{g}_i,  \mathbf{g}_i^{+}, \mathbf{G}_{i}^{-}) \\%\underbrace{\mathbf{g}_{i}^{--}, \mathbf{g}_{i,1}^-,...,\mathbf{g}_{i,n^{-}}^-}_{\mathbf{G}_{i}^{-}} ) \\
            %&=L(\mathbf{g}_i,  \mathbf{g}_i^{+}, \mathbf{G}_{i}^{-}) \\
        %&= - \log \frac{ e^{\textrm{sim}(\mathbf{g}_i,\mathbf{g_{i}^{+}})}}{ e^{\textrm{sim}(\mathbf{g}_i,\mathbf{g_{i}^{+}})} + e^{\textrm{sim}(\mathbf{g}_i,\mathbf{g_{i}^{--}})} + \sum_{j=1}^{n^i}e^{\textrm{sim}(\mathbf{g}_i,\mathbf{g}_j)}}
        &= - \log \frac{ e^{\textrm{sim}(\mathbf{g}_i,\mathbf{g}_{i}^{+})}}{ e^{\textrm{sim}(\mathbf{g}_i,\mathbf{g}_{i}^{+})} + \sum_{\mathbf{g}\in\mathbf{G}_{i}^{-}}e^{\textrm{sim}(\mathbf{g}_i,\mathbf{g})}}.
\end{align}
%In addition to the conventional cross entropy (CE) loss for training a logistic regression, we optimise the joint loss function:
In Appendix~\ref{appendix:sim_loss}, we compare different similarity metrics and loss functions.

\subsection{Retrieval-based KNN classifier}
% Similar to HateCLIPper, we use conventional logistic regression for classification.
%To assess the expressiveness and discrimination capability of the trained joint embedding space, 
In addition to logistic classifier, we introduce a retrieval-based KNN majority voting classifier which relies on the inherent discrimination capability of the trained joint embedding space.
Only when the trained embedding space successfully splits hateful and benign examples will majority voting achieve reasonable performance. 
The KNN classifier is suitable for real services in the constantly evolving landscape of online hateful memes as the the retrieval database can be extended without the need to retrain the system.
In Section~\ref{sec:results_knn}, we show that our proposed KNN classifier generalizes well to unseen data without additional training.

%For each test meme, we retrieve memes located in close proximity within the embedding space and utilise weighted majority voting to predict whether it is hateful or not. 
%We use the cosine similarity as the weight for the majority voting.
For a test meme $t$, we retrieve $K$ memes located in close proximity within the embedding space from the retrieval vector database $\mathbf{G}$ (see Eq.~\ref{eq:database}). We keep a record of the retrieved memes' labels $y_k$ and similarity scores $s_k=\text{sim}(g_k, g_t)$ with the test meme $t$, 
%\begin{equation}
%    \{(y_k,s_k:\text{sim}(g_k, g_t))\}_{k=1}^{K},
%\end{equation}
where $g_t$ is the embedding vector of the test meme $t$.
We perform similarity-weighted majority voting to obtain the prediction:
\begin{equation}
    \hat{y}'_t = \sigma(\sum_{k=1}^K\Bar{y}_k \cdot s_k),
\end{equation}
where $\sigma(\cdot)$ is the sigmoid function and 
\begin{equation}
    \Bar{y}_k:=
    \begin{cases}
        1  &\text{if } y_k= 1\\
        -1 &\text{if } y_k=0
    \end{cases}.
    \label{eq:indicator_function}
\end{equation}
%\[
%y'_k=2y_k - 1
%\]

%During inference, we encode training examples to form a retrieval database $\mathbf{G}$(refer to Eq.~\ref{eq:database}).
% We retain the label for each sample.
%\begin{equation}
%    \mathbf{G}=\{(\mathbf{g}_j, y_j)\}_{j=1}^N
%\end{equation}
%where $\mathbf{g}_j = \mathcal{F}(I_j,T_j)$.
%For an unseen test example, we encode its image and text pair to form vector $\mathbf{g}_x$. Then we retrieve $K$ most similar training examples with Faiss nearest neighbour search. 
%Finally, the labels of retrieved training examples determine whether the test sample is hateful by majority voting.
% Benign samples contributes negative logits, yielding small values of $\hat{y}_t'$ which corresponds to benign prediction. Same-label samples contributes positive logits. 
We conduct experiments in Sec.~\ref{sec:results_knn} to show that applying RGCL leads to much better performance with retrieval-based KNN inference than using only the cross-entropy loss. 
% the majority voting score that is computed based on the top K similar examples' label is used to determine whether the test sample $x$ is hateful or not. 
\section{RGCL experiments}
%\subsection{Dataset}
%We evaluate the performance of the system on the hateful memes challenge dataset (HMC) \cite{KielaFBHMC2020}. To make a fair comparison, we adopt the evaluation metrics commonly used in existing hateful meme classification studies \cite{KumarHateClip2022, caoPromptHate2022, KielaFBHMC2020}: Area Under the Receiver Operating Characteristic Curve (AUROC) and Accuracy (Acc). Then, to test the generalizing capability, we evaluate the performance of the HarMeme\cite{pramanickCovidMeme2021} \cite{harm}datasets that contain memes related to COVID-19. The HarMeme datasets contains 
%Initially, we implement and develop our system on the HatefulMemes dataset \cite{KielaFBHMC2020}. Subsequently, to test the generalising capability, we evaluate the performance of the HarMeme \cite{pramanickCovidMeme2021}. 
We primarily evaluate the performance of RGCL on the \textbf{HatefulMemes} dataset \cite{KielaFBHMC2020} and the \textbf{HarMeme} dataset \cite{pramanickCovidMeme2021}. %HatefulMemes dataset is released by the Hateful Memes Challenge competition \cite{KielaFBHMC2020}.
The HarMeme dataset consists of COVID-19-related harmful memes collected from Twitter. In Section~\ref{sec:abl_generalize}, we evaluate three additional datasets to show the generalizability of RGCL beyond hateful meme classification. The dataset statistics are shown in Appendix~\ref{appendix:data_stats}. 

To make a fair comparison, we adopt the evaluation metrics used in previous literature \cite{KumarHateClip2022, caoPromptHate2022, KielaFBHMC2020} for HatefulMemes and HarMeme: Area Under the Receiver Operating Characteristic Curve (AUC) and Accuracy (Acc).

%We train the system on the training split, refine it on the development splits and report the final results on the test set.
%We tune the hyperparameters on the development split.
%We develop our system on the HatefulMemes and use the same hyperparameter settings for training on HarMeme. 
The experiment setup, including the statistical significance tests, and hyperparameter settings are detailed in Appendices~\ref{appendix:exp_setup} and \ref{appendix:hyperparam}.

%\subsection{Experiment results} 
\subsection{Comparing RGCL with baseline systems}
Table~\ref{tab:results_HMC} presents the experimental results with logistic regression.
RGCL is compared to a range of baseline models including OD-based models, LMMs, and CLIP-based systems. On the \textbf{HatefulMemes} dataset, RGCL obtains an AUC of $87.0\%$ and an accuracy of $78.8\%$, outperforming all baseline systems, including the 200 times larger Flamingo-80B.
\\
\textbf{OD-based models}\\ ERNIE-Vil \cite{ErnieViL2020}, UNITER \cite{Uniter2019} and OSCAR \cite{li2020oscar} performs similarly with AUC scores of around $79\%$.
%Furthermore, systems built on the original CLIP significantly improve the performance over CLIP, except for the MOMENTA \cite{PramanickMomenta2021}. Momenta achieves a modest AUC of 69.2, which is comparably lower than CLIP's AUC of 79.8. MOMENTA was initially developed on the HarMeme \cite{PramanickMomenta2021} dataset with a complicated modality and feature fusion mechanism. Its approach might overfit to the HarMeme dataset, leading to suboptimal performance on the HatefulMemes dataset. 
\\\textbf{LMMs}\\
Flamingo-80B \cite{Flamingo22} is the previous state-of-the-art model for HatefulMemes, with an AUC of $86.6\%$. %Thus, we reproduce LLaVA to understand performance on state-of-the-art open source LMM.}. 
%Our model attains a slightly higher AUC of $87.0\%$, setting a new benchmark result on the HatefulMemes dataset despite having 16,000 times fewer trainable parameters and 200 times fewer total parameters. 
We also %transform HatefulMemes into instruction following data and 
fine-tune LLaVA \cite{LiuLLAVA2023} with the procedure in Appendix~\ref{appendix:llava}. LLaVA achieves $77.3\%$ accuracy and $85.3\%$ AUC, performing worse than the much larger Flamingo, but better than OD-based models. 
\\\textbf{CLIP-based systems}\\
PromptHate \cite{caoPromptHate2022} and HateCLIPper \cite{KumarHateClip2022}, built on top of CLIP \cite{clip2021}, outperform both the original CLIP and OD-based models.
HateCLIPper achieves an AUC of $85.5\%$, surpassing the original CLIP (79.8\% AUC) but falling short of Flamingo-80B (86.6\% AUC). 
%Our approach, utilising HateCLIPper's modelling, further enhances performance with an AUC of $87.0\%$. Notably, our system's accuracy also improves over HateCLIPper by nearly $3\%$. 
%HateCLIPper, trained using our proposed \textbf{RGCL}, 
Our system, utilising HateCLIPper's modelling, improves over HateCLIPper by nearly $3\%$ in accuracy, reaching $78.8\%$. For the AUC score, our system achieves $87.0\%$, surpassing the previous state-of-the-art Flamingo-80B.

%It is noteworthy that our system demands significantly fewer resources compared to LLaVA. Instruction fine-tuning on the HatefulMemes dataset requires 4 A100-80GB of 12 hours of runtime, while our system requires a single 24GB GPU for a half-hour runtime. For inference, LLaVA requires around 30GB of memory, while our system operates efficiently with only 2GB. 

%Table~\ref{tab:results_Harmeme} shows the result on the \textbf{HarMeme} dataset. %Being a less popular dataset, there are no previous results available for large multimodal models, we similarly fine-tune LLaVA to obtain the results on LMM.
%Notably, in contrast to the degradation over CLIP as observed in the HatefulMemes dataset, MOMENTA demonstrates an improvement on the HarMeme dataset with an accuracy of $80.5\%$, surpassing the original CLIP model's accuracy of $76.7\%$.
For \textbf{HarMeme}, RGCL obtained an accuracy of $87\%$, outperforming HateCLIPper with an accuracy of $84.8\%$, PromptHate with an accuracy of $84.5\%$ and LLaVA with an accuracy of $83.3\%$. Our system's state-of-the-art performance on the HarMeme dataset further emphasises RGCL's robustness and generalisation capacity to different types of hateful memes.
\newcounter{mpFootnoteValueSaver}
\begin{table}[htb]
\small
\centering
\setcounter{mpFootnoteValueSaver}{\value{footnote}}
\begin{tabularx}{0.48\textwidth}{X|ll|ll}
\toprule 
 \multicolumn{3}{r}{\textbf{HatefulMemes}} & \multicolumn{2}{r}{\textbf{HarMeme}} \\
 \multicolumn{1}{X}{Model}                     & \textbf{AUC} & \multicolumn{1}{l}{ \textbf{Acc}.}     & \textbf{AUC} & \textbf{Acc}.  \\ 
\midrule
\multicolumn{3}{l}{\textit{~~~~~Object Detector based models}} \\
 \midrule
ERNIE-Vil  & 79.7 & 72.7 & - & -\\
UNITER & 79.1 & 70.5 & - & -\\
OSCAR  & 78.7 & 73.4 & - & -\\
\midrule
\multicolumn{3}{l}{\textit{~~~~~Fine-tuned Large Multimodal Models}}                             \\ \midrule
Flamingo-80B\footnotemark & 86.6 & - & - & -\\
LLaVA (Vicuna-13B) & 85.3 & 77.3  & 90.8 & 83.3  \\
\midrule
%\citet{RonHMC1st2020} & 84.5 & 73.2 \\
 \multicolumn{3}{l}{\textit{~~~~~Systems based on CLIP}} \\
 \midrule
CLIP  & 79.8 & 72.0 & 82.6 & 76.7\\
MOMENTA & 69.2 & 61.3 & 86.3 & 80.5\\
PromptHate &81.5 & 73.0 & 90.9 & 84.5\\ 
HateCLIPper\footnotemark & 85.5 & 76.0 & 89.7  & 84.8    \\
HateCLIPper \textbf{w/ RGCL} & \textbf{87.0} & \textbf{78.8}  & \textbf{91.8} & \textbf{87.0} \\
\bottomrule
\end{tabularx}
\caption{Comparing RGCL with baseline systems. Best performance is in \textbf{bold}.}
\label{tab:results_HMC}
\end{table}

\subsection{Performance with retrieval-based KNN classifier}
\label{sec:results_knn}
%We present the results on the \textbf{HatefulMemes} dataset using KNN-based majority voting classifier.
Online hate speech is constantly evolving, and it is not practical to keep retraining the detection system. We demonstrate that our system can effectively transfer to the unseen domain of hateful memes without retraining.  

We train HateCLIPper with and without RGCL using the HarMeme dataset and evaluate on the HatefulMemes dataset. We report the performance of the KNN classifier when using the HarMeme and HatefulMemes dataset as the retrieval database in Table~\ref{tab:results_retrieval} ({\Romannum{2}}) and ({\Romannum{3}}) respectively. We only use the training set as the retrieval database to avoid label leaking.

We compare our method with state-of-the-art LMMs, including Flamingo \cite{Flamingo22}, Lens \cite{BerriosLens2023}, Instruct-BLIP \cite{Ouyang_InstructGPT_2022} and LLaVA \cite{LiuLLAVA2023} as shown in Table~\ref{tab:results_retrieval} ({\Romannum{1}}). We report the zero-shot performance of these LMMs to replicate the scenario when the model predicts the unseen domain of hateful memes. To ensure a fair comparison, we report the performance of LLaVA fine-tuned on the HarMeme to align with RGCL's setting in Table~\ref{tab:results_retrieval} ({\Romannum{2}}) and ({\Romannum{3}}). 

Lastly, we also report the performance of our methods when trained and evaluated on HatefulMemes in Table~\ref{tab:results_retrieval} ({\Romannum{4}}).

\stepcounter{mpFootnoteValueSaver}%
\footnotetext[\value{mpFootnoteValueSaver}]{{Since Flamingo is not open-sourced, we are unable to obtain accuracy.}}
\stepcounter{mpFootnoteValueSaver}%
\footnotetext[\value{mpFootnoteValueSaver}]{Reproduced with HateCLIPper's code base.}
\begin{table}[!hbtp]
\small

\centering
\begin{tabularx}{0.495\textwidth}{Xll}
\toprule
 Model                & \textbf{AUC} & \textbf{Acc}. \\ 
\midrule
\multicolumn{3}{l}{\textit{~~~~~(\textbf{\Romannum{1}}) Zero shot based on Large Multimodal Models}}                             \\ \midrule
%\multicolumn{2}{l}{Flamingo-9B} & 57.0 & -\\
{Flamingo-80B} & 46.4 & - \\
{Lens \textit{(Flan-T5 11B)}}  & 59.4 & - \\
{InstructBLIP \textit{(Flan-T5 11B)}}  & 54.1 & - \\
{InstructBLIP \textit{(Vicuna 13B)}}  & 57.5 & -  \\ 
{LLaVA \textit{(Vicuna 13B)}}  & 57.9 & 54.8 \\
\textit{~~~~fine-tuned on HarMeme}  & 56.3 & 54.3 \\ 
%LLaVA (Vicuna-13B) & Failed & Failed \\
\midrule

\multicolumn{3}{l}{\textit{~~~~~(\textbf{\Romannum{2}}) Train and retrieve on HarMeme }}                             \\ \midrule
%Training set & Retrieval set & \textbf{AUC} & \textbf{Acc}. \\
HateCLIPper &  55.8 & 51.9 \\
\textit{~~~~LR instead of KNN} & \textit{52.4} & \textit{49.5} \\
HateCLIPper \textbf{w/ RGCL}&  \textbf{60.0 \footnotesize{\textit{(\textcolor{olive}{+4.2})}}} & \textbf{57.2 \footnotesize{\textit{(\textcolor{olive}{+5.3})}}} \\
\textit{~~~~LR instead of KNN} & \textit{59.4} \footnotesize{\textbf{\textit{(\textcolor{olive}{+7.0})}}} & \textit{50.9 } \footnotesize{\textbf{\textit{(\textcolor{olive}{+1.4})}}} \\
\midrule
\multicolumn{3}{l}{\textit{~~~~~(\textbf{\Romannum{3}}) Train on HarMeme, retrieve on HatefulMemes}}                             \\ \midrule
HateCLIPper &  54.4 & 50.3 \\
HateCLIPper \textbf{w/ RGCL}&   \textbf{66.6 \footnotesize{\textit{(\textcolor{olive}{+12.2})}}} & \textbf{59.9 \footnotesize{\textit{(\textcolor{olive}{+9.6})}}} \\
%HatefulMemes & HatefulMemes& 78.5 & 73.6  \\
%HarMeme & HarMeme  & 54.8 & 52.2 \\
%HarMeme & HatefulMemes & 53.6 & 50.9 \\

\midrule
\multicolumn{3}{l}{\textit{~~~~~(\textbf{\Romannum{4}}) Train and retrieve on HatefulMemes}}                             \\ \midrule
%Training set & Retrieval set & \textbf{AUC} & \textbf{Acc}. \\
%\midrule

HateCLIPper &  84.6 & 73.3  \\
HateCLIPper \textbf{w/ RGCL}&  \textbf{86.7 \footnotesize{\textit{(\textcolor{olive}{+2.1})}}} & \textbf{78.3 \footnotesize{\textit{(\textcolor{olive}{+5.0})}}}  \\
%HarMeme & HarMeme  & 59.8 & 57.1 \\
%HarMeme & HatefulMemes & 66.6 & 59.1 \\

\bottomrule
\end{tabularx}
\caption{Retrieval-based KNN classifier results on HatefulMemes. LR refers to logistic regression.}
\label{tab:results_retrieval}
\end{table}

%\\
%\textbf{LMMs}\\ 

(\textbf{\Romannum{1}}) We report LMMs with diverse backbone language models, ranging from Flan-T5 \cite{Chung_FLAN_2022} and the more recent Vicuna \cite{vicuna2023}.
Among these models, Lens with Flan-T5XXL 11B performs the best, achieving an AUC of $59.4\%$.
%Instruct-BLIP with the Vicuna 13B as the language model achieves an AUC of $57.5\%$.Following closely, Flamingo-9B achieves an AUC of $57.0\%$. Notably, the larger 80B version of Flamingo attains a significantly lower AUC of only $46.4\%$.
%\\
%\textbf{Retrieval-based KNN Classifier} 
%\\
When LLaVA is fine-tuned on the HarMeme dataset and evaluated on the HatefulMemes dataset, its performance does not improve beyond its zero-shot performance. Its accuracy drops from $54.8\%$ in zero-shot to $54.3\%$ in fine-tuned. These findings indicate that the fine-tuned LLaVA struggles to generalise effectively to diverse domains of hateful memes.  

%For retrieval-based KNN classifier, We report models' performance on the \textbf{HatefulMemes} dataset when trained on the HarMeme dataset.
(\textbf{\Romannum{2}})  When using the \textbf{HarMeme} as the retrieval database, our system achieves an AUC of $60.0\%$, surpassing both the baseline HateCLIPper's AUC of $55.8\%$ and the best LMM's zero-shot AUC score. 

Additionally, we provide the results of using logistic regression (LR) as an alternative to the KNN classifier, both with and without RGCL training, when systems trained on HarMeme are tested on HatefulMemes. The performance of logistic regression consistently falls short of the KNN classifier. Logistic regression with RGCL training achieves an AUC of $59.4\%$, outperforming the HateCLIPper's baseline by $7\%$. Note that the logistic regression does not the retrieval of examples.

(\textbf{\Romannum{3}}) When using \textbf{HatefulMemes} as the retrieval database,
the HateCLIPper's performance degrades, suggesting its embedding space lacks generalizing capability to different domains of hateful memes.
RGCL boosts the AUC to $66.6\%$, outperforming the baseline by a large margin of $12.2\%$. RGCL achieves an accuracy of $59.9\%$, surpassing the baseline by $9.6\%$. RGCL's AUC and accuracy score also surpass the zero-shot LMMs.
%Our result, with 20 times fewer parameters, outperforms the best large multimodal model's zero-shot performance.
%Incorporating the training data of HatefulMemes into the retrieval database further enhances our model's AUC score to $66.6\%$, highlighting the improved adaptability of our system to new meme domains. This adaptability is crucial for real-world hate speech detection systems, given the constantly evolving landscape of online hate speech.

(\textbf{\Romannum{4}}) %The AUC is lower than the logistic classifier as shown in Table~\ref{tab:results_HMC} due to the KNN inference does not output raw logits. The AUC is lower ... since the KNN inference only outputs the majority voted label. Logistic regression shows better AUC than the KNN due to the fact that 
When our system is trained and evaluated on the HatefulMemes dataset (the same system from Table~\ref{tab:results_HMC}), the KNN classifier obtains $86.7\%$ AUC and $78.3\%$ accuracy. %, which are close to the logistic classifier's AUC of $87.0\%$ AUC and accuracy of $78.8\%$.
These scores also surpass all baseline systems including fine-tuned LMMs in Table~\ref{tab:results_HMC}. %RGCL's accuracy outperformance HateCLIPper by a margin of $4.6\%$.

%\subsection{Ablation Study}
\subsection{Effects of incorporating pseudo-gold positive and hard negative examples}
\label{sec:abl_exp}
In Table~\ref{tab:ablation_examples}, we report a comparative analysis by examining performance when specific examples are excluded during the training process. 

When we omit the pseudo-gold positive examples, only in-batch positive examples are incorporated during the training. This results in an accuracy degradation of $1.5\%$. 
Hard positive examples, same-label samples with high similarity scores, are commonly used in contrastive learning literature. In our case,
when incorporating hard positive examples rather than pseudo-gold positive examples, the training becomes unstable and results in divergence.
%When the hard negative examples are excluded, leaving only in-batch negative samples, the overall quality of negative samples declines.
%Similar degradation is observed, where the AUC and accuracy degrade to $86.1\%$ and $77.1\%$ respectively. 

When the hard negative examples are excluded, leaving only in-batch negative samples, the performance degrades $1.7\%$ for accuracy.
When removing both types of examples, there is more performance degradation. Both the pseudo-gold positive examples and the hard negative examples are needed for accurately classifying hateful memes. 

When excluding the in-batch negative examples, training becomes unstable and fail to converge, which is consistent with previous findings in  \citep{Henderson2017effiNLPinbatchnegative}.
%Specifically, there is a decrease of $0.6\%$ and $0.7\%$ in AUC when omitting hard negative and pseudo-gold positive examples, respectively.  Furthermore, the accuracy metric experiences a more substantial reduction, with drops of $1.7\%$ and $1.5\%$ for the respective cases. Notably, the combined exclusion of both the hard negative and pseudo-gold examples results in a marked decrease in performance. This discrepancy is apparent in the AUC score, which experiences a substantial drop when compared to our baseline. The AUC only matches with HateCLIPper's performance, as indicated in Table~\ref{tab:results_HMC}. Additionally, the accuracy of $76.8\%$ slightly outperforms HateCLIPper's accuracy of $76.0\%$.

%Furthermore, we train RGCL with more than one example for each scenario. %The inclusion of two hard negative examples leads to substantial performance deterioration, with corresponding drops of $0.8\%$ and $1.5\%$ in AUC and accuracy. Training with two pseudo-gold positive examples yields a slight decline in performance, resulting in a $0.2\%$ decrease in AUC and a $0.3\%$ decrease in accuracy. 
\begin{table}[htb]
\small
\centering
\begin{tabularx}{0.45\textwidth}{Xll}
\toprule
 Model                     & \textbf{AUC} & \textbf{Acc}.  \\ 
\midrule
Baseline RGCL & \textbf{87.0} & \textbf{78.8}\\ \midrule
%w/o in-batch negative & 86.6 & 78.6 \\
%\midrule
%\multicolumn{3}{l}{\textit{~~~~~Training without some of the examples}}                             \\ \midrule
w/o Pseudo-Gold positive & 86.0 & 77.3 \\
w/o Hard negative & 86.1 & 77.1 \\
w/o Hard negative and Pseudo-gold positive & 85.5 &  76.8\\
%\midrule
%\multicolumn{3}{l}{\textit{~~~~~Training with more than one example }}                             \\ \midrule
%w/ 2 Hard negative & 85.9 & 77.3 \\
%w/ 2 Pseudo-Gold positive & 86.6 & 78.5 \\
\bottomrule
\end{tabularx}
\caption{Ablation study on omitting Hard negative and/or Pseudo-Gold positive examples on the HatefulMemes}
\label{tab:ablation_examples}
\end{table}

\subsection{Effects of different VL Encoder}
\label{sec:abl_encoder}
We ablate the performance when incorporating RGCL on various VL encoders. As shown in Table~\ref{tab:ablation_encoder}, we experiment with various encoders in the CLIP family: the original CLIP \cite{clip2021}, OPENCLIP \cite{ilharco_gabriel_OPENCLIP2021,schuhmann2022laionbopenclip,cherti2023reproducibleopenclip}, and AltCLIP \cite{chen2022altclip}. 
Our method boosts the performance of all these variants of CLIP by around $3\%$. 

To verify that our method does not depend on the CLIP architecture, we carry out experiments with ALIGN\footnote{ALIGN only open-sourced the base model which is less capable than the larger CLIP-based models.} \cite{Jia2021ALIGN}. As shown in Table~\ref{tab:ablation_encoder}, RGCL enhances the AUC score by a margin of $4.4\%$ over the baseline ALIGN model.

\begin{table}[htb]
\small
\centering
\begin{tabularx}{0.42\textwidth}{Xll}
\toprule
 Model                     & \textbf{AUC} & \textbf{Acc}.  \\ 
\midrule
%\multicolumn{3}{l}{\textit{~~~~~HateCLIPper}}   \\ \midrule
%HateCLIPper & 85.5 &76.0 \\
%HateCLIPper w/ RGCL & 87.0 \textbf{\textit{(\textcolor{olive}{+1.5})}} & 78.8 \textbf{\textit{(\textcolor{olive}{+2.8})}} \\
%\textit{gain} & +1.2 & +2.8 \\
%w/o in-batch negative & 86.6 & 78.6 \\
%\midrule
%\multicolumn{3}{l}{\textit{~~~~~Original CLIP}}                             \\ \midrule
CLIP & 79.8 & 72.0\\
CLIP w/ RGCL & 83.8 \textbf{\textit{(\textcolor{olive}{+4.0})}} & 75.8 \textbf{\textit{(\textcolor{olive}{+3.8})}}\\
%\textit{gain} &+4.0 & +3.8\\
\midrule
%\multicolumn{3}{l}{\textit{~~~~~OpenCLIP }}                             \\ \midrule
OpenCLIP  & 82.9 & 71.7 \\
OpenCLIP w/ RGCL & 84.1 \textbf{\textit{(\textcolor{olive}{+1.2})}} & 75.1 \textbf{\textit{(\textcolor{olive}{+3.4})}}\\
\midrule
%\multicolumn{3}{l}{\textit{~~~~~AltCLIP }}                             \\ \midrule
AltCLIP & 83.4 & 74.1\\
AltCLIP w/ RGCL & 86.5 \textbf{\textit{(\textcolor{olive}{+3.1})}} & 76.8 \textbf{\textit{(\textcolor{olive}{+2.7})}}\\
\midrule
ALIGN &  73.2& 66.8  \\
ALIGN w/ RGCL & 77.6 \textbf{\textit{(\textcolor{olive}{+4.4})}} & 68.9 \textbf{\textit{(\textcolor{olive}{+2.1})}} \\
\bottomrule
\end{tabularx}
\caption{Ablation study on various VL encoders on the HatefulMemes dataset}
\label{tab:ablation_encoder}
\end{table}

\subsection{Effects of dense/sparse retrieval}
%Pseudo-gold positive examples and hard negative examples can be obtained by either dynamic dense retrieval or static sparse retrieval during training.
We compare the commonly used sparse retrieval to our proposed dynamic dense retrieval for obtaining contrastive learning examples.
%Image-to-text transform methods like object detection is required in sparse retrieval since it can only be applied on text. 
%To perform sparse retrieval, we carry out image-to-text transformation using object detection. 
%Here, we compare the performance of dense retrieval and sparse retrieval during training process to obtain the retr. 
We detail our approach for sparse retrieval in Appendix~\ref{appendix:sparse_retrieval}.   

As shown in Table~\ref{tab:ablation_sparse}, %we compare baseline dense retrieval with different sparse retrieval settings. 
using a variable number of objects in object detection performs the best in sparse retrieval. However, the accuracy degrades by $0.7\%$ compared to the dense retrieval baseline. When using a fixed number of objects in object detection, the performance degrades even more. Our proposed dynamic dense retrieval obtains better performance than the commonly used sparse retrieval methods.
\begin{table}[htb]
\small
\centering
\begin{tabularx}{0.4\textwidth}{Xll}
\toprule
 Model             & \textbf{AUC} & \textbf{Acc}.  \\ 
\midrule
%\multicolumn{3}{l}{\textit{~~~~~Training with Dense Retrieval}}                             \\ \midrule
Baseline \textit{with Dense Retrieval} & 87.0 & 78.8\\
%w/o in-batch negative & 86.6 & 78.6 \\
\midrule
%\multicolumn{3}{l}{\textit{~~~~~Training with Sparse Retrieval}}                             \\ \midrule
%\textit{~~~~with Sparse Retrieval} \\
w/ Variable No. of objects & 87.0 & 78.1 \\
w/ 72 objects  & 86.1 & 77.1 \\
w/ 50 objects & 85.9 & 78.6  \\ 
\bottomrule
\end{tabularx}
\caption{Ablation study of Dense retrieval and Sparse retrieval to obtain pseudo-gold positive examples and hard negative examples on the HatefulMemes dataset}
\label{tab:ablation_sparse}
\end{table}
%As shown in Table~\ref{tab:results_HMC}'s last row, the sparse retrieval method achieves the same AUC of $87.0\%$ with the dense retrieval, but with a slightly lower accuracy of $78.1\%$. 

\subsection{Effects of Retrieval-Guided Contrastive Learning Loss}
As shown in Eq.~\ref{eq:RGCLL}, the mixing ratio between RGCLL and the CE loss is 1:1 by default. In Table~\ref{tab:mix_ratio}, we compare the different mixing ratios between the two loss functions. We observe a significant performance improvement whenever RGCLL is included. For simplicity, we maintain a 1:1 mixing ratio.
Notably, in the absence of cross-entropy loss, we identified several examples where models with RGCL fail but models without RGCL succeed. Conversely, inclusion of cross-entropy loss eliminates such discrepancies. 
\begin{table}[h]
\centering
\small
\begin{tabularx}{0.325\textwidth}{cll}
\toprule
\textbf{RGCLL:CE} & \textbf{Acc.} & \textbf{AUC} \\
\midrule
0:1 & 76.0 & 85.5 \\
0.5:1 & 78.5 & 86.8 \\
1:1   & 78.8 & \textbf{87.0} \\
2:1   & \textbf{79.1} & 86.9 \\
4:1   & 78.6 & 86.9 \\
1:0   & 79.0 & 86.5 \\
\bottomrule
\end{tabularx}
\caption{Ablation study of different mixing ratios for the two type of loss functions on the HatefulMemes dataset}
\label{tab:mix_ratio}
\end{table}

\subsection{Effects of RGCL on different Meme Classification tasks}
\label{sec:abl_generalize}
To demonstrate RGCL's versatility beyond hateful meme classification, we assess its efficacy on three additional datasets: MultiOFF~\cite{suryawanshi-etal-2020-MultiOFF}, Harm-P~\cite{PramanickMomenta2021}, and Memotion7K~\cite{Sharma_Memotion_2020}. These datasets originally used the F1 score as their evaluation metric; we also include Accuracy. We train a separate model for each of the datasets following the procedures detailed in Appendices~\ref{appendix:exp_setup} and \ref{appendix:hyperparam}.   Table~\ref{tab:results_generalize} shows the results for CLIP with and without RGCL training on these three datasets\footnote{Since CLIP surpasses almost all other prior published systems on these datasets, we do not include prior results in the comparison.}. 
%RGCL generalizes well on all these datasets, emphasizing its usefulness across various multimodal classification tasks.
%which tackles offensive messages on Instagram, politically harmful memes on Twitter and multimodal sentiment analysis on memes respectively. 

\textbf{MultiOFF} contains memes related to the 2016 U.S. presidential election sourced from social media sites, such as Twitter and Instagram. The memes are labeled as non-offensive and offensive. MultiOFF is a relatively small dataset, containing less than 500 training examples. RGCL outperforms the baseline by a significant margin of $4.7\%$ in accuracy. RGCL yields consistent gain even with relatively small datasets.

\textbf{Harm-P} contains harmful and harmless memes on US politics sourced from social media sites. RGCL shows more than 2\% gain in both accuracy and F1 score over the baseline system.

\textbf{Memotion7K}, designed for multi-task meme emotion analysis, includes annotations for humor, sarcasm, offensiveness, and motivation. RGCL shows improvement over baseline across all four emotion classification tasks with an average gain of more than 3\% on both accuracy and F1 scores. These results highlight RGCL's capability for improving emotion detection.

\begin{table}[htb]
\small
\centering
\begin{tabularx}{0.49\textwidth}{X|cc|cc}
\toprule 
\multicolumn{3}{r}{\textbf{w/o RGCL}} & \multicolumn{2}{c}{\textbf{w/ RGCL}} \\                     \multicolumn{1}{l}{Dataset} & \textbf{Acc}. & \multicolumn{1}{l}{ \textbf{F1}}     & \textbf{Acc}. & \textbf{F1}  \\ 
\midrule
MultiOFF  & 62.4 & 54.8 & 67.1 \textbf{\textit{(\textcolor{olive}{+4.7})}}  & 58.1 \textbf{\textit{(\textcolor{olive}{+3.3})}} \\
\midrule
{Harm-P} &  87.6 & 86.9 & 89.9 \textbf{\textit{(\textcolor{olive}{+2.3})}}  & 89.5  \textbf{\textit{(\textcolor{olive}{+2.6})}}  \\
\midrule
%\citet{RonHMC1st2020} & 84.5 & 73.2 \\
Memotion7K &  &  &  & \\
 %\midrule

\textit{~~~~{-Humour}} &  73.0 & 83.8  & 76.3 \textbf{\textit{(\textcolor{olive}{+3.3})}} & 86.6 \textbf{\textit{(\textcolor{olive}{+2.4})}} \\
\textit{~~~~{-Sarcasm}} & 75.1 & 85.6  & 77.3 \textbf{\textit{(\textcolor{olive}{+2.2})}}   & 87.2  \textbf{\textit{(\textcolor{olive}{+1.6})}}   \\
\textit{~~~~{-Offensive}} & 72.8& 83.5 & 77.6 \textbf{\textit{(\textcolor{olive}{+4.8})}} & 87.4 \textbf{\textit{(\textcolor{olive}{+3.9})}}  \\
\textit{~~~~{-Motivation}} & 59.6 & 72.6 & 62.4 \textbf{\textit{(\textcolor{olive}{+2.8})}} & 76.8 \textbf{\textit{(\textcolor{olive}{+4.2})}}\\ 
\textit{{Average}} &70.1 & 81.4 & 73.4 \textbf{\textit{(\textcolor{olive}{+3.3})}} & 84.5 \textbf{\textit{(\textcolor{olive}{+3.1})}}    \\

\bottomrule
\end{tabularx}
\caption{The performance of CLIP with and without RGCL training on different meme classification tasks}
\label{tab:results_generalize}
\end{table}
% add a note that CLIP is better than the previous published results 
\section{Case Analysis}
\label{sec:qualitative}
We now analyze how RGCL improves relative to baseline systems on confounding memes.
\subsection{Quantitative analysis}
From the 500 validation samples of HatefulMemes, we annotated 101 examples and picked 24 confounder memes. On this confounder subset, HateCLIPper without RGCL obtains an accuracy of 66.7\%, while RGCL significantly boosts the accuracy to 83.3\%. These results show that RGCL improves the classification of challenging confounder memes, which exhibit differences in either the image or text. 

Next, we analyze how RGCL improves the classification through examples of confounder memes from the subset.
\subsection{Qualitative analysis}
%We show confounding examples in Table~\ref{tab:counfounder_visualisation}. 
In Table~\ref{tab:counfounder_visualisation}, we demonstrate how RGCL addresses the classification errors associated with confounder memes. Our approach significantly reduces the similarity scores between anchor memes and confounder memes. This shows that RGCL effectively learns a hatefulness-aware embedding space, placing the meme within the embedding space with a comprehensive hateful understanding derived from both vision and language components. By aligning semantically similar memes closer and pushing apart dissimilar ones in the embedding space, RGCL enhances classification accuracy.

\begin{table*}[hbtp]
\centering
\small
(a)\\ \vspace{1pt}
\begin{tabularx}{\textwidth}{Xccc}
\toprule
 & Anchor memes & Image confounders & Text confounders \\ 
 Ground truth labels & Hateful & Benign & Benign \\
\midrule
 Meme & \includegraphics[valign=c, width=0.24\textwidth]{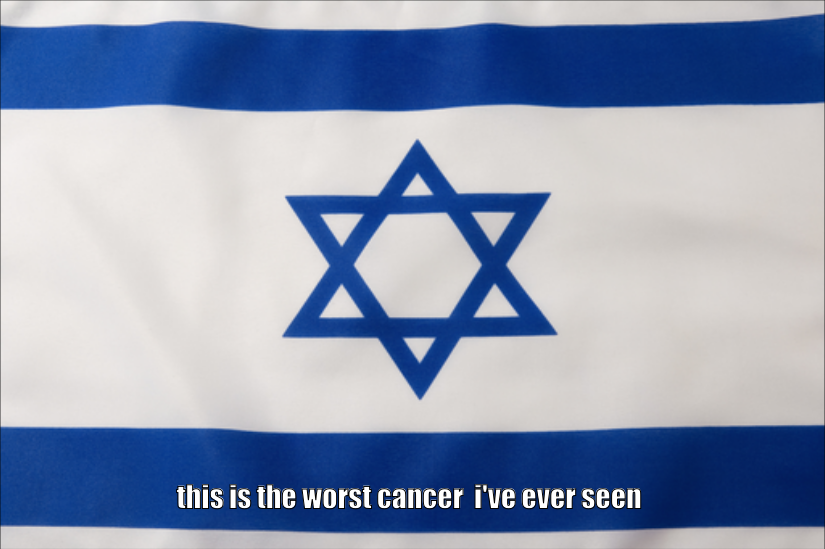} & \includegraphics[valign=c,width=0.24\textwidth]{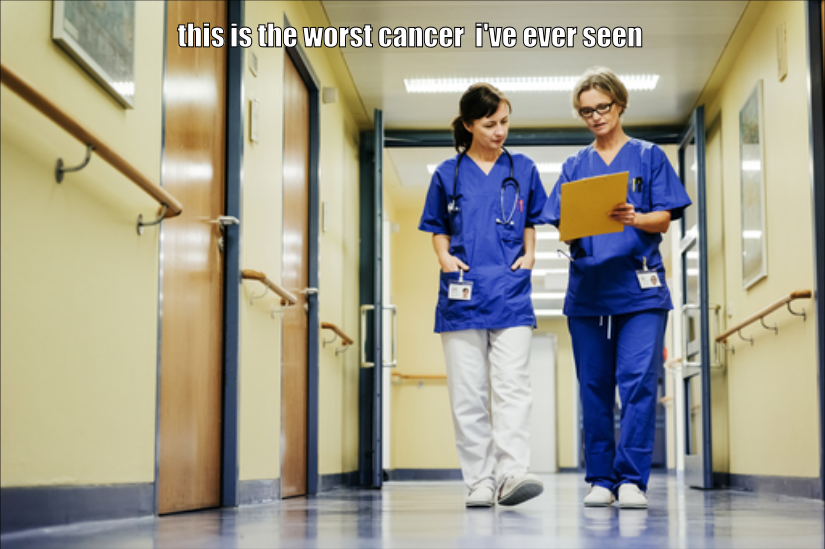} & \includegraphics[valign=c, width=0.24\textwidth]{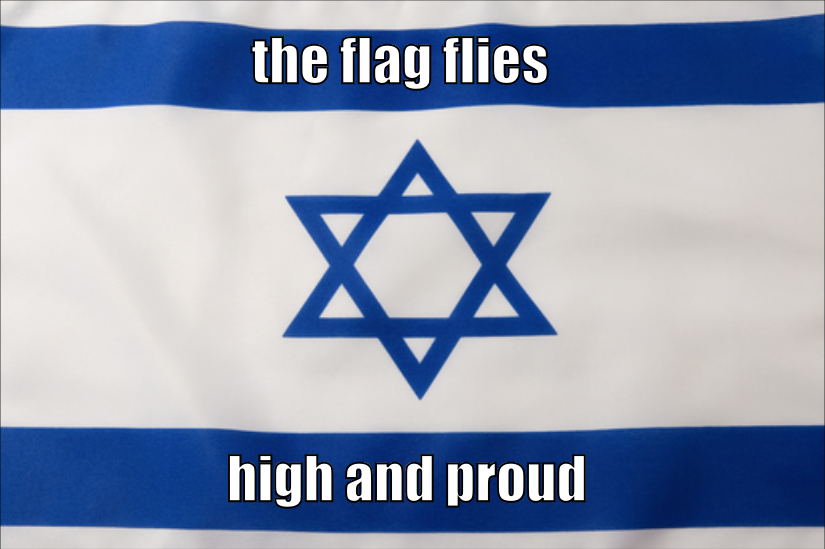} \\
\midrule
%Splits: & Test & Train & Train \\
%\midrule
 \multicolumn{4}{l}{\textit{~~~~~HateCLIPper}} \\
 \midrule
Probability & 0.454 & 0.000 & 0.001 \\  
Prediction & \textcolor{red}{Benign \xmark}  & Benign & Benign \\  
\multicolumn{2}{l}{Similarity with anchor \hspace{7em} -} & 0.702 & 0.733 \\
\midrule
 \multicolumn{4}{l}{\textit{~~~~~HateCLIPper w/ RGCL (Ours)}} \\
 \midrule
Probability & 0.999 & 0.000  & 0.000 \\
Prediction  & \textbf{\textcolor{green}{Hateful \cmark}} & Benign  & Benign \\
%\midrule
\multicolumn{2}{l}{Similarity with anchor \hspace{7em} -} & \textbf{-0.751} & \textbf{-0.571} \\
%\bottomrule
\midrule
\end{tabularx}
(b)\\
\vspace{1pt}
\begin{tabularx}{\textwidth}{Xccc}
%\toprule
% & Anchor Meme & Image Confounder & Text Confounder \\ 
\midrule
 Meme & \includegraphics[valign=c, width=0.235\textwidth]{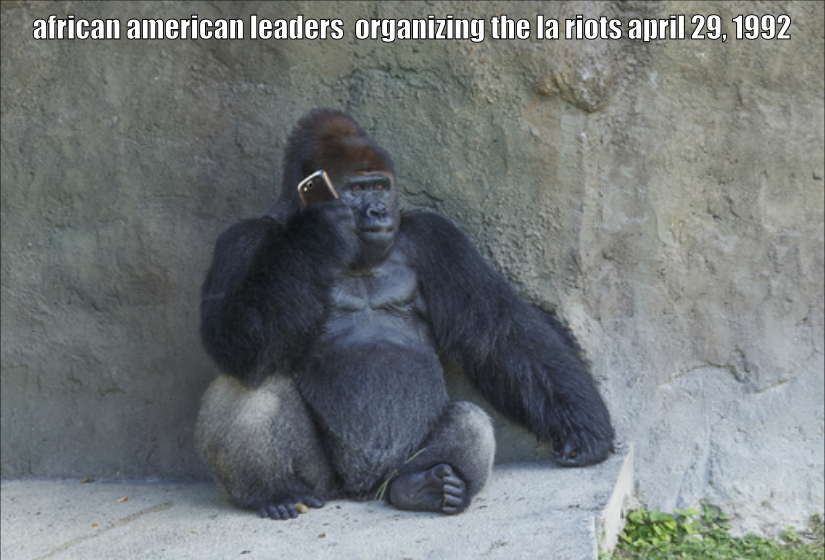} & \includegraphics[valign=c, width=0.235\textwidth]{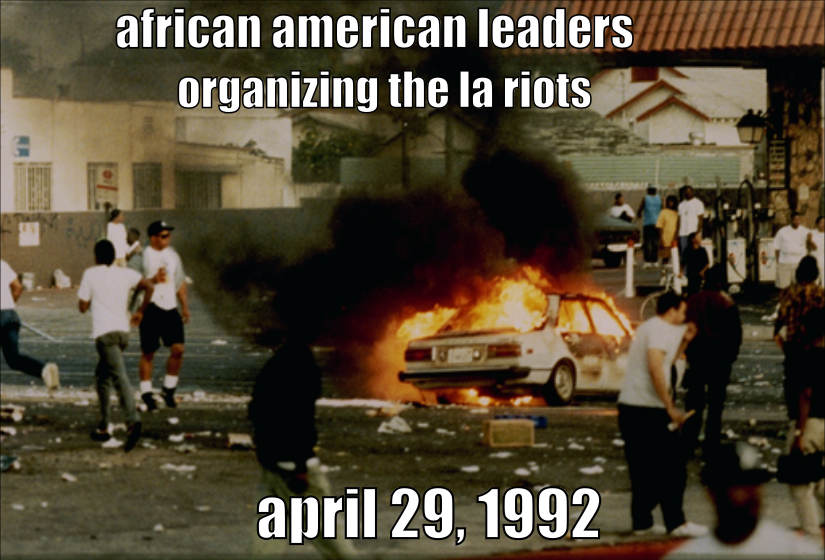} & \includegraphics[valign=c, width=0.235\textwidth]{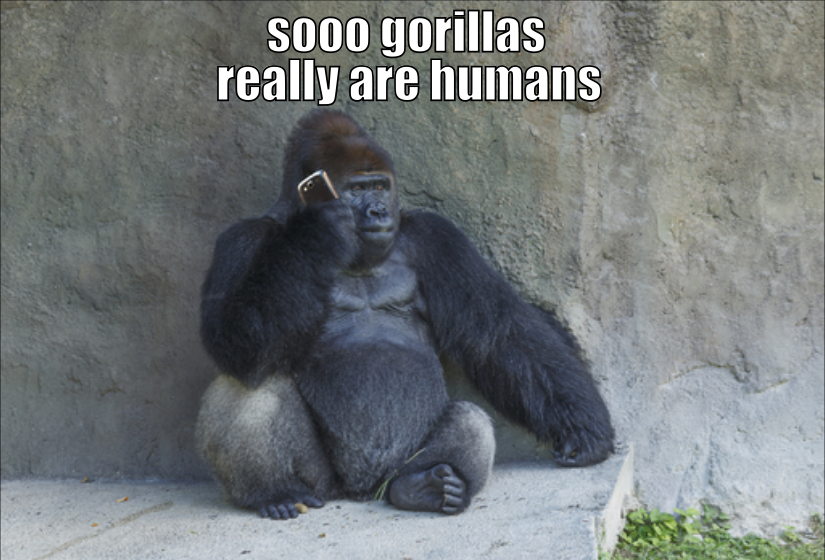} \\
 \midrule
%Split: & Test & Train & Train \\
%\midrule
%Labels & Hateful & Benign & Benign \\
%\midrule
 \multicolumn{4}{l}{\textit{~~~~~HateCLIPper}} \\
 \midrule
Probability  & 0.038 & 0.000 & 0.001 \\  

Prediction & \textcolor{red}{Benign \xmark} & Benign & Benign \\  
\multicolumn{2}{l}{Similarity with anchor \hspace{7em} -} & 0.898 & 0.913 \\
\midrule
 \multicolumn{4}{l}{\textit{~~~~~HateCLIPper w/ RGCL (Ours)}} \\
\midrule
Probability& 1.00 & 0.000  & 0.000 \\
Prediction & \textbf{\textcolor{green}{Hateful \cmark}} & Benign  & Benign \\
\multicolumn{2}{l}{Similarity with anchor \hspace{7em} -} &\textbf{ -0.803} & \textbf{-0.769} \\
\midrule
\end{tabularx}
(c)\\
\vspace{1pt}
\begin{tabularx}{\textwidth}{Xccc}
%\toprule
% & Anchor Meme & Image Confounder & Text Confounder \\ 
\midrule
 Meme & \includegraphics[valign=c, width=0.235\textwidth, height=1.8cm]{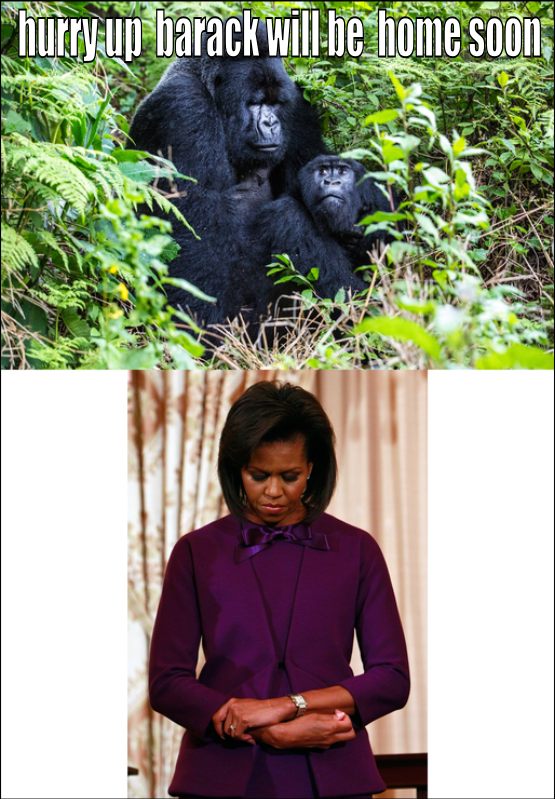} & \includegraphics[valign=c, width=0.235\textwidth, height=1.8cm]{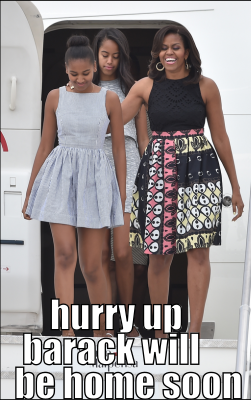} & \includegraphics[valign=c, width=0.235\textwidth, height=1.8cm]{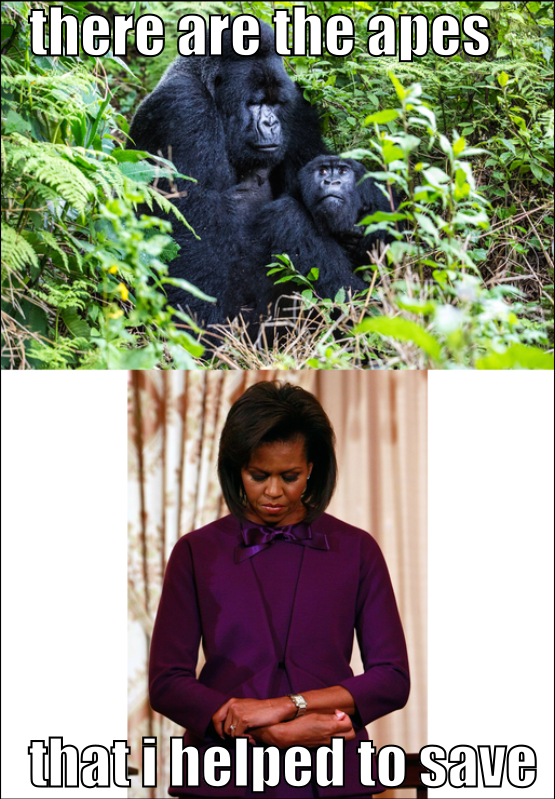} \\
 \midrule
%Split: & Test & Train & Train \\
%\midrule
%Labels & Hateful & Benign & Benign \\
%\midrule
 \multicolumn{4}{l}{\textit{~~~~~HateCLIPper}} \\
 \midrule
Probability  & 0.385 & 0.001 & 0.005 \\  

Prediction & \textcolor{red}{Benign \xmark} & Benign & Benign \\  
\multicolumn{2}{l}{Similarity with anchor \hspace{7em} -} & 0.869 & 0.781 \\
\midrule
 \multicolumn{4}{l}{\textit{~~~~~HateCLIPper w/ RGCL (Ours)}} \\
\midrule
Probability& 0.996 & 0.000  & 0.000 \\
Prediction & \textbf{\textcolor{green}{Hateful \cmark}} & Benign  & Benign \\
\multicolumn{2}{l}{Similarity with anchor \hspace{7em} -} &\textbf{ -0.980} & \textbf{-0.998} \\
\bottomrule
\end{tabularx}
\caption[Visualisation for the Confounder memes in the HatefulMemes dataset]{Visualisation for the confounder memes in the HatefulMemes dataset. We present triplets of memes including the hateful anchor memes, the benign image confounders and the benign text confounders. We show the output hateful probability and predictions from HateCLIPper and our RGCL system. We provide the cosine similarity score between the anchor meme and its corresponding confounder meme.}
\label{tab:counfounder_visualisation}
\end{table*}
\section{Conclusion}
We introduce Retrieval-Guided Contrastive Learning to enhance any VL encoder in addressing challenges in distinguishing confounding memes. Our approach uses novel auxiliary loss with dynamically retrieved examples and significantly improves contextual understanding. 
Achieving an AUC score of $87.0\%$ on the HatefulMemes dataset, our system outperforms prior state-of-the-art models. Our approach also transfers to different tasks, emphasizing its usefulness across diverse meme domains.

\section*{Limitation}
%From the perspective of annotation of 
Hate speech can be defined by different terminologies, such as online harassment, online aggression, cyberbullying, or harmful speech. United Nations Strategy and Plan of Action on Hate Speech stated that the definition of hateful could be controversial and disputed \cite{united_nations_2020}. 
Additionally, according to the UK's Online Harms White Paper, harms could be insufficiently defined \cite{uk_parliament_2022}. We use the definition of hate speech from the two datasets: HatefulMemes \cite{KielaFBHMC2020} and HarMeme \cite{pramanickCovidMeme2021}.
These datasets adopt Facebook's definition of hate speech \footnote{\href{https://transparency.fb.com/en-gb/policies/community-standards/hate-speech/}{https://transparency.fb.com/en-gb/policies/community-standards/hate-speech/}} to strike a balance between reducing harm and preserving freedom of speech. Tackling the complex issue of how to define hate speech will require a cooperative effort by stakeholders, including governmental policy makers, academic scholars, the United Nations Human Rights Council, and social media companies. 
We align our research with the ongoing process of defining the hate speech problem and will continue to integrate new datasets, as they become available.

%On the technical side, current state-of-the-art systems still perform far from satisfactory. For example, Table~\ref{tab:error_visualisation} shows a trio of memes from the HatefulMemes dataset, adopting a structure similar to Table~\ref{tab:counfounder_visualisation}. The Anchor meme portrays a person with an exaggeratedly elongated nose with a caption of "when your jewish friend smells a stash of coins in public". This meme carries implicit offensiveness towards the Jewish community. Our method correctly categorizes it as hateful, marking an improvement over the HateCLIPper model's performance.  In the case of the image confounder, the meme substitutes the image with one depicting a person discovering a dirty can in public, displaying a disgusted facial expression. The combination of text and image renders this meme benign. However, neither of the two systems successfully identifies this meme as hateful. 
%Table~\ref{tab:error_visualisation} shows a trio of memes from the HatefulMemes dataset, adopting a structure similar to Table~\ref{tab:counfounder_visualisation}. Both models misclassify the middle meme.
In examining the error cases of our system, we find that the system is unable to recognize subtle facial expressions. This can be improved by using a more powerful vision encoder to enhance image understanding. We leave this to future work. 
%This limitation might arise from the models' inability to comprehend facial expressions, which remains a constraint of our approach. Such challenges could potentially be addressed with a more advanced vision encoder.

\section*{Ethical Statement}
\paragraph{Reproducibility.} We present the detailed experiment setups and hyperparameter settings in Appendices~\ref{appendix:exp_setup} and \ref{appendix:hyperparam}. The source code will be released upon publication.
\paragraph{Usage of Datasets.}
The HatefulMemes, HarMeme, MultiOFF, and Harm-P datasets were curated and designed to help fight online hate speech for research purposes only.
Throughout the research, we strictly follow the terms of use set by their authors.

\paragraph{Societal benefits.}
Hate speech detection systems like RGCL contribute significantly to reducing online hate speech, promoting safer digital environments, and aiding in protecting human content moderators. These positive impacts, we believe, are substantial and crucial in the broader context of online communication and safety.
\paragraph{Intended use.}
We intend to enforce strict access controls upon the model release. The model will only be shared with researchers after signing the terms of use. We will clearly state that the system is intended for the detection and prevention of hateful speech. We will specify that it should not be used for any purposes that promote, condone, or encourage hate speech or harmful content. 
\paragraph{Implementation consideration.}
Because our system is based on retrieving examples, multiple retrieval sets reflect different cultural sensitivities that can be applied in reality. Our architecture is well suited to addressing the problem of cultural differences or subjective topics without retraining.
However, the annotation of datasets in handling cultural differences or subjective topics needs to be take into consideration before any deployment of systems. The factors need to be considered includes the data curation guidelines, bias of the annotators, and the limited definition of hate speech. 
\paragraph{Misuse Potential.} Our proposed system does not induce biases. However, training the system on HatefulMemes or HarMeme may cause unintentional biases towards certain individuals, groups, and entities \cite{PramanickMomenta2021}. To counteract potential unfair moderation stemming from dataset-induced biases, incorporating human moderation is necessary.

\paragraph{Environmental Impact}
Training large-scale Transformer-based models requires a lot of computations on GPUs/TPUs, which contributes to global warming. However, this is a bit less of an issue for our system, since we only fine-tune small components of vision-language models. Our system can be trained under 30 minutes on a single GPU. The fine-tining takes far less time compared to LMMs. Moreover, as our model is relatively small, the inference cost is much less compared to LMMs.

\section*{Acknowledgments}
Jingbiao Mei is supported by Cambridge Commonwealth, European and International Trust for the undertaking of the PhD in Engineering at the University of Cambridge.

Jinghong Chen is supported by the Warwick Postgraduate Studentship from Christ’s College and the Huawei Hisilicon Studentship for the undertaking of the PhD in Engineering at the University of Cambridge.

Weizhe Lin is supported by a Research Studentship funded by Toyota Motor Europe (RG92562(24020)) for the undertaking of the PhD in Engineering at the University of Cambridge.

Prof. Bill Byrne holds concurrent appointments as a Professor of Information Engineering at Cambridge University and as an Amazon Scholar.  This publication describes work performed at Cambridge University and is not associated with Amazon.

We would also like to thank all the reviewers for their knowledgeable reviews.

\bibliography{anthology,custom}
\bibliographystyle{acl_natbib}

\appendix
\label{sec:appendix}

\section{Experiment Setup}
\label{appendix:exp_setup}
A work station equipped with NVIDIA RTX 3090 and AMD 5900X was used for the experiments. \texttt{PyTorch 2.0.1}, \texttt{CUDA 11.8}, and \texttt{Python 3.10.12} were used for implementing the experiments. HuggingFace transformer library \cite{HuggingFace_trans_2019} was used for implementing the pretrained CLIP encoder \cite{clip2021}. Faiss \cite{johnson_Faiss2019billion} vector similarity search library with version \texttt{faiss-gpu 1.7.2} was used to perform dense retrieval. Sparse retrieval was performed with \texttt{rank-bm25 0.2.2} \footnote{\href{https://github.com/dorianbrown/rank\_bm25}{https://github.com/dorianbrown/rank\_bm25}}. All the reported metrics were computed by \texttt{TorchMetrics 1.0.1}. %\footnote{\href{https://torchmetrics.readthedocs.io}{https://torchmetrics.readthedocs.io}. TorchMetrics version 1.0 and above has resolved the bug related to computing the micro F1 score.}. 
For LLaVA \cite{LiuLLAVA2023}, we fine-tuned the model on a system with 4 A100-80GB. The runtime was 4 hours on the HatefulMemes and 3 hours on the HarMeme. The details for fine-tuniung is covered in Appendix~\ref{appendix:llava}. All the metrics were reported based on the mean of three runs with different seeds. Due to the limited space in Table~\ref{tab:results_HMC}, we provide more details for our main results here.  HateCLIPper with RGCL obtained an accuracy of $78.77 \pm 0.25$ and an AUC of $86.95 \pm 0.21$ on HatefulMemes.

\section{Hyperparameter}
\label{appendix:hyperparam}
The default hyperparameter for all the models are shown in Table~\ref{tab:hyperparameters}. The modelling hyperparameter is based on HateCLIPper's setting \cite{KumarHateClip2022} for a fair comparison. For vision and language modality fusion, we perform element-wise product between the vision embeddings and language embeddings. This is known as align-fusion in HateCLIPper~\cite{KumarHateClip2022}. The hyperparameters associated with retrieval-guided contrastive learning are manually tuned with respect to the evaluation metric on the development set. With this configuration of hyperparameter, the number of trainable parameters is about 5 million and training takes around 30 minutes.
\begin{table}[h]
\small
\centering
\caption{Default hyperparameter values for the modelling and Retrieval-Guided Contrastive Learning (\textbf{RGCL})}
\label{tab:hyperparameters}
\begin{tabular}{lc}
\toprule
Modelling hyperparameter & Value \\
\midrule
Image size & 336 \\
Pretrained CLIP model & ViT-L-Patch/14 \\
Projection dimension of MLP & 1024 \\
Number of layers in the MLP & 3 \\
Optimizer & AdamW \\
Maximum epochs & 30 \\
Batch size & 64 \\
Learning rate & 0.0001 \\
Weight decay & 0.0001 \\
Gradient clip value & 0.1 \\
Modality fusion & Element-wise product \\
\midrule 
\midrule
RGCL hyperparameter  & Value \\
\midrule
\# hard negative examples & 1 \\
\# pseudo-gold positive examples & 1 \\
Similarity metric & Cosine similarity \\
Loss function & NLL \\
Top-K for retrieval based inference & 10\\ 

\bottomrule
\end{tabular}
\end{table}

\section{HateCLIPper's Architecture}
\label{appendix:hateclipper}
For the $i^{\textrm{th}}$ image and text pair $(I_i, T_i)$, HateCLIPper obtains the feature embeddings $f_{I}$ and $f_{T}$ \footnote{Dropped subscript $i$ for simplicity} with the pretrained CLIP vision and language encoders. To facilitate the learning of task-specific features, distinct trainable projection layers are employed after the extracted feature vectors to obtain projected features $f_{I}'$ and $f_{T}'$. $f_{I}'$ and $f_{T}'$ are vectors of dimension $n$, which is a hyperparameter to tune. These trainable \textbf{projection layers} consist of a feedforward layer followed by a dropout layer. These feature vectors undergo explicit cross-modal interaction via \textbf{Hadamard product} i.e., element-wise multiplication. This fusion process is referred to "\textbf{align-fusion}" within the HateCLIPper framework. After the align-fusion, a series of \textbf{Pre-Output layers} are employed, comprising multiple feedforward layers incorporating activation functions and dropout layers. These layers are applied to the image and text representation $f_{I}'$ and $f_{T}'$ to obtain the final embedding vector $\mathbf{g}_i$. The number of Pre-Output layers is a hyperparameter to tune. We shorthand this process of obtaining the joint embedding vector with $\mathcal{F}(\cdot,\cdot)$ for simplification as denoted in Eq.~\ref{eq:Encoder}.

\section{Dataset details and statistics}
\label{appendix:data_stats}
Table~\ref{tab:dataset_stats} shows the data split for the HatefulMemes and HarMeme datasets. Note that HarMeme is first introduced in \citealt{pramanickCovidMeme2021}, however, in \citealt{PramanickMomenta2021}, HarMeme had been renamed to Harm-C. Following the notation of previous works \cite{caoPromptHate2022}, we use its original name HarMeme in this paper.
The memes in HarMeme are labeled with three classes: \textit{very harmful, partially harmful}, and \textit{harmless}. Following previous work \cite{caoPromptHate2022, PramanickMomenta2021}, we combine the very harmful and partially harmful memes into hateful memes and regard harmless memes as benign memes.

\begin{table}[h]
\small
\centering
\caption{Statistical summary of HatefulMemes and HarMeme datasets}
\begin{tabular}{c|cc|cc}
\toprule
Datasets  & \multicolumn{2}{c}{Train}  & \multicolumn{2}{c}{Test}  \\
& \#Benign & \#Hate & \#Benign & \#Hate  \\
\midrule
HatefulMemes &   5450 & 3050 & 500 & 500              \\
\midrule
HarMeme  & 1949 & 1064 & 230 & 124             \\

\bottomrule
\end{tabular}%
\label{tab:dataset_stats}
\end{table}
In addition to hateful memes classification, we also evaluate the MultiOFF, Harm-P and Memotion7K datasets. Table~\ref{tab:dataset_stats_generalize} shows the dataset statistics.
\begin{table}[h]
\small
\centering
\caption{Statistical summary of MultiOFF, Harm-P and Memotion7K datasets. Neg. for Negative, Pos. for Positive.}
\begin{tabularx}{\linewidth}{X|cc|cc}
\toprule
Datasets  & \multicolumn{2}{c}{Train}  & \multicolumn{2}{c}{Test}  \\
& \#Neg. & \#Pos. & \#Neg. & \#Pos.  \\
\midrule
MultiOFF(Offensive)&   258 & 187 & 58 & 91              \\
\midrule
Harm-P(Harmful)  & 1534 & 1486 & 182 & 173             \\
\midrule 
Memotion7K \\
\textit{~~~~{-Humour}} & 1651 & 5342& 445& 1433\\
\textit{~~~~{-Sarcasm}} & 1544 & 5449 & 421& 1457 \\
\textit{~~~~{-Offensive}} & 2713 & 4280 & 707& 1171 \\
\textit{~~~~{-Motivation}} & 4526 & 2467  & 1188&690  \\ 
\bottomrule
\end{tabularx}%
\label{tab:dataset_stats_generalize}
\end{table}

To access the Facebook HatefulMemes dataset, one must follow the license from Facebook\footnote{\href{https://hatefulmemeschallenge.com/\#download}{https://hatefulmemeschallenge.com/\#download}}.HarMeme and Harm-P is distributed for research purpose only, without a license for commercial use. 
MultiOFF is licensed under CC-BY-NC. Memotion7K has no specific license mentioned.

\section{LLaVA experiments}
\label{appendix:llava}
For fine-tuning LLaVA~\cite{LiuLLAVA2023}, we follow the original hyperparameters setting\footnote{\href{https://github.com/haotian-liu/LLaVA}{https://github.com/haotian-liu/LLaVA}} for fine-tuning on downstream tasks. For the prompt format, we follow InstructBLIP~\cite{DaiInstructBLIP2023}. For computing the AUC and accuracy metrics, we also follow InstructBLIP's procedure.

\section{Ablation study on numbers of retrieved examples }
We experiment with using more than one hard negative and pseudo-gold positive gold examples in training.

The inclusion of more than one example for both types of examples causes the performance to degrade.
This phenomenon aligns with recent findings in the literature, as \citet{dpr2020} reported that the incorporation of multiple hard negative examples does not necessarily enhance performance in passage retrieval.
\begin{table}[htb]
\small
\caption{Ablation study on omitting and using two Hard negative and/or Pseudo-Gold positive examples on the HatefulMemes}
\label{tab:ablation_examples_two}
\centering
\begin{tabularx}{0.4\textwidth}{Xll}
\toprule
 Model                     & \textbf{AUC} & \textbf{Acc}.  \\ 
\midrule
Baseline RGCL & \textbf{87.0} & \textbf{78.8}\\ \midrule
%w/o in-batch negative & 86.6 & 78.6 \\
%\midrule
%\multicolumn{3}{l}{\textit{~~~~~Training without some of the examples}}                             \\ \midrule
%w/o Hard negative & 86.1 & 77.1 \\
%w/o Pseudo-Gold positive & 86.0 & 77.3 \\
%w/o Hard negative and Pseudo-gold positive & 85.5 &  76.8\\
%\midrule
%\multicolumn{3}{l}{\textit{~~~~~Training with more than one example }}                             \\ \midrule
w/ 2 Hard negative & 85.9 & 77.3 \\
w/ 4 Hard negative & 85.7 & 76.0 \\
w/ 2 Pseudo-Gold positive & 86.6 & 78.5 \\
w/ 4 Pseudo-Gold positive & 86.3 & 77.4 \\
\bottomrule
\end{tabularx}
\end{table}
\section{Ablation study on loss function and similarity metrics}
\label{appendix:sim_loss}
Inner product (IP) and Euclidean L2 distance are also commonly used as similarity measures. Since Euclidean distance (L2) is a distance metric, we take its negative to serve as a measure of similarity. We tested these alternatives and found cosine similarity performs slightly better as shown in Table~\ref{tab:ablation_loss_sim}. 

Additionally, another popular loss function for ranking is triplet loss \cite{Chechik_LargeScale_ImageSim_Rank2009, Schroff_FaceNet_2015} which compares a positive example with a negative example for an anchor meme. Our results in Table~\ref{tab:ablation_loss_sim} suggest that using triplet loss performs comparably to the default NLL loss.
\begin{table}[htbp]
\small
\caption{Ablation study on the loss function and similarity metrics on the HatefulMemes dataset. Similarity metrics include cosine similarity, inner product and negative squared L2.}
\label{tab:ablation_loss_sim}
\centering
\begin{tabularx}{0.4\textwidth}{XXll}
\toprule
Loss & Similarity           & \textbf{AUC} & \textbf{Acc}.  \\ 
\midrule
\multirow{3}{*}{NLL} & Cosine & \textbf{87.0} & \textbf{78.8} \\
 & Inner Product & 86.1 & 78.2 \\
 & L2 & 85.7 & 76.6  \\
%w/o in-batch negative & 86.6 & 78.6 \\
\midrule
%\multicolumn{3}{l}{\textit{~~~~~Training with other loss function}}                             \\ \midrule
\multirow{3}{*}{Triplet} & Cosine & 86.7 & 78.7 \\
 & Inner Product & 86.1 & 78.2 \\
 & L2 &  85.7 & 76.8 \\

%\midrule
%\multicolumn{3}{l}{\textit{~~~~~Training with other similarity metrics}}                             \\ \midrule
%w/ Inner Product & 86.1& 78.2  \\
%w/ Negative Squared L2 & 85.8 & 76.8  \\
\bottomrule
\end{tabularx}
\end{table}

\section{Sparse retrieval}
\label{appendix:sparse_retrieval}
We use VinVL object detector \cite{vinVL2021} to obtain the region-of-interest object prediction and its corresponding attributes. 

After obtaining these text-based image features, we concatenate these text with the overlaid caption from the meme to perform the sparse retrieval. We use BM-25 \cite{Robertson_BM25_2009} to perform sparse retrieval.
%We use dense retrieval to obtain pseudo-gold positive and hard negative examples to avoid an additional pipeline of object detection. However, in previous literature like dense passage retrieval \cite{dpr2020}, sparse retrieval methods like BM-25 \cite{Robertson_BM25_2009} are used to obtain hard negative examples to avoid dynamically encoding the vector retrieval database. Here, we also ablate the performance of our system when incorporating sparse retrieval to obtain the retrieved examples. \howard{TODO}
For variable number of object predictions, we set a region-of-interest bounding box detection threshold of $0.2$, a minimum of 10 bounding boxes, and a maximum of 100 bounding boxes, consistent with the default settings of the VinVL.

\end{document}